\documentclass[11pt]{article}
\usepackage{geometry}
\usepackage{coling2020}
\usepackage{times}
\usepackage{url}
\usepackage{latexsym}
\usepackage{microtype}
\usepackage{todonotes}
\usepackage{graphicx}
\usepackage{hyperref}
\usepackage{subcaption}
\usepackage{mwe}
\usepackage[utf8]{inputenc} % arxiv admin

\colingfinalcopy % Uncomment this line for all SemEval submissions

\title{ANDES at SemEval-2020 Task 12: A jointly-trained BERT multilingual model for offensive language detection}

\author{Juan Manuel Pérez\thanks{Authors contributed equally}\\
  ICC, CONICET \\
  Universidad de Buenos Aires \\
  {\small {\tt jmperez@dc.uba.ar}} \\\And
  Aymé Arango\footnotemark[1] \\
  University of Chile \\
  IMFD, Chile\\
  {\small {\tt aarango@dcc.uchile.cl}}  \\\And
  Franco M. Luque \\
  CONICET \\
  Universidad Nacional de Córdoba \\
  {\small {\tt francolq@famaf.unc.edu.ar}} \\}

\date{\today}

\begin{document}
\maketitle
\begin{abstract}
This paper describes our participation in SemEval-2020 Task 12: Multilingual Offensive Language Detection. We jointly-trained a single model by fine-tuning Multilingual BERT to tackle the task across all the proposed languages: English, Danish, Turkish, Greek and Arabic.
Our single model had competitive results, with a performance close to top-performing systems in spite of sharing the same parameters across all languages.
Zero-shot and few-shot experiments were also conducted to analyze the transference performance among these languages. We make our code public for further research\footnote{\url{https://github.com/finiteautomata/offenseval2020}}.
\end{abstract}

\section{Introduction}

\blfootnote{
    %
    % for review submission
    %
    % % final paper: en-us version 
    
    \hspace{0.0cm}  % space normally used by the marker
    This work is licensed under a Creative Commons 
    Attribution 4.0 International License.
    License details:
    \url{http://creativecommons.org/licenses/by/4.0/}.
}

Offensive language, hate speech, cyberbullying, and abusive language detection are topics that have summoned a lot of interest in the last years, particularly due to the necessity in Social Media to stop --or at least, to diminish-- this omnipresent phenomenon. Not only the social implications are behind this, but also practical implications for companies: recently, some large advertisers have removed their presence from Social Media as they consider that platforms full of ``divisiveness and hate speech'' do not give value to their companies \cite{HateSpeechGuardian2020}.

The differences between these categories are loosely defined. \newcite{waseem2017understanding} propose using two axes to understand the particularities of each discourse: whether the abusive language is directed to a specific individual or to a generalized group, and whether the offending discourse is explicit or implicit.

Even when online platforms prohibit behavior that crosses the line into abuse,\footnote{ https://help.twitter.com/en/safety-and-security/offensive-tweets-and-content} these rules are frequently violated. Automated moderation algorithms are necessary to perform a faster and even better user-generated-content moderation or to serve as a tool that helps human moderators to reduce the volume of offensive content still present in online platforms. To address this problem, offensive language detection is usually thought of as a binary classification problem in which the input is a post (tweet, message, comment, etc.) and the output is the classification of whether the post is offensive or not.

Several features of offensive language make it complex to detect. The \textit{offenders} could intentionally hide offensive words by substituting letters with special characters or numbers. Content with irony or sarcasm could be harmful even when all words are polite, and vice versa. The real intention behind speech is sometimes difficult to detect even for humans.

According to \newcite{DBLP:conf/websci/ChatzakouKBCSV17}, in the case of Twitter, each tweet provides a fairly limited context. Therefore, an offensive post may be identified as unoffensive if the context is not taken into account.
Despite several published proposals on detecting offensive language, hate speech, and other related concepts, there is no consolidated or effective strategy that could be applied to different languages and domains.

Building a quality labeled dataset is an expensive task in time and effort.
%and remains to be a problem for the researchers in the area.
Most available datasets are in English, and often datasets are not published due to privacy concerns, making research on offensive language detection difficult for other languages.
%, and also they are often not published 
%
The generalization power of learning models would allow us to transfer knowledge to languages with poor resources. %With that purpose, we constructed a single model for tackling the task across all the languages proposed.

In this work, we describe our participation in SemEval 2020 Task 12: OffensEval, an offensive language detection task in five different languages. We propose a single multilingual system based on Multilingual BERT \cite{devlin2018bert} to jointly address the offensive language detection. We trained our system using all the available training data and evaluated its performance for each of the languages instead of adjusting language-specific models. Our single model performs fairly well, achieving performances comparable to the winning teams for each language. To further explore the multilingual dimension of BERT, we analyze zero-shot and few-shot capabilities in a cross-lingual setting for the task in question. We make our code public for further research.

\section{Background}
The detection of offensive language, cyberbullying and hate speech are tasks that are closely connected and often confused \cite{malmasi2018challenges}. Several machine learning models addressing hate speech or offensive language detection have been proposed in the last years; in particular \textit{deep learning}  models \cite{CNN1,CNN2,badjatiya2017deep,DBLP:conf/ecir/AgrawalA18,bisht2020detection,gertner2019mitre,perez2019atalaya} have increased their popularity among researchers on this task. Despite the growing interest in the area, the models are usually trained and evaluated inside very specific English datasets, and their generalizability to other contexts or languages is still a challenge. Morever, building these datasets is difficult \cite{waseem2016you} and achieved performances are often overestimated \cite{arango2019hate,wiegand2019detection}.

There are only a few studies addressing multilingual detection of these subtypes of abusive language in the related literature. In these, authors proposed single systems that can be used to classify data in different languages. Some of the common features used in this kind of models are multilingual word representations such as MUSE \cite{conneau2017word} or Multilingual BERT \cite{DBLP:conf/naacl/DevlinCLT19}. Other authors combined word embeddings with tweet-level features \cite{corazza2020multilingual} or linguistic features \cite{DBLP:conf/semeval/BenitoAI19}.

In most cases, the multilingual models are trained and tested independently for each language and do not combine different languages in a single evaluation. An exception is the approach proposed by \newcite{DBLP:conf/semeval/BojkovskyP19}, where the authors trained deep neural network architectures with a concatenation of English and Spanish datasets to classify data in both languages.

Although Multilingual BERT models have been tested as end-to-end solutions for several tasks, they have not been widely explored for offensive language detection. \newcite{pires2019multilingual} tested the zero-shot capability of BERT for transferring knowledge from one language to another in  \textit{named entity recognition} and \textit{part of speech tagging} tasks obtaining high performing results.

\section{Data}
The dataset used in this work is described thoroughly in \newcite{zampieri-etal-2020-semeval}. We decided to replace the distantly-supervised English training dataset\cite{rosenthal2020} presented in this task by the OLID dataset of \newcite{zampierietal2019}. This is because the focus of this work is the multilingual capabilities of BERT, and the datasets of the languages added to this task (Danish\cite{sigurbergsson2020offensive}, Greek\cite{pitenis2020}, Arabic\cite{mubarak2020arabic} and Turkish\cite{coltekikin2020}) resemble more the OLID dataset in that they were manually annotated.

\begin{table}[h!]
	\centering
	\begin{tabular}{lrrr}
		Language & Train & Dev & Test\\
		\hline
		English  & 13240  & 860  & 3887  \\
		Greek    & 6994   & 1749 & 1544  \\
		Danish   & 2368   & 592  & 329   \\
		Arabic   & 6839   & 1000 & 2000  \\
		Turkish  & 25021  & 6256 & 3528  \\
        \hline
  	\end{tabular}%

  	\caption{Size of the train, development and test datasets used in this work.}
		\label{tab:data}
\end{table}%            

\section{System Overview}
Our model is a fine-tuned version of Multilingual BERT \cite{devlin2018bert}. This architecture (which has become the state of the art for most NLP tasks) consists of a stack of transformer blocks \cite{vaswani2017attention} pretrained in two tasks: masked language model (also known as Cloze task) and next sentence prediction. For the downstream task an output layer is added and the model is fine-tuned with very low learning rates.
%These pretrained blocks are then added an output layer for the downstream task, and finally fine-tuned with very low learning rates.

Multilingual BERT shares the same training as single-language BERT but using a concatenated dataset of 104 languages, and has demonstrated to have surprising cross-lingual capabilities, even among languages that do not share scripts \cite{pires2019multilingual}.

The implementation used in this work is the pretrained multilingual BERT-base model from the HuggingFace library\cite{wolf2019huggingface}.
This model consists of 12 transformer blocks, 12 self-attention heads, and a hidden layer size of 768. On the top of it, we added a linear layer and applied a sigmoid function to the outputs.

We trained the model for 10 epochs with a batch size of 32 using a dropout probability of 0.1, setting the initial learning rate at $5 * 10^{-5}$ and binary cross entropy as the loss function. Adam with linear warm-up of 10\% of the steps was used to optimize the loss.

% Key algorithms and modeling decisions in your system; resources used beyond the provided training data; challenging aspects of the task and how your system addresses them. This may require multiple pages and several subsections, and should allow the reader to mostly reimplement your system’s algorithms.

% Use equations and pseudocode if they help convey your original design decisions, as well as explaining them in English. If you are using a widely popular model/algorithm like logistic regression, an LSTM, or stochastic gradient descent, a citation will suffice—you do not need to spell out all the mathematical details.

% Give an example if possible to describe concretely the stages of your algorithm.

% If you have multiple systems/configurations, delineate them clearly.

% This is likely to be the longest section of your paper.

\section{Experimental Setup}
We tested several experimental configurations using the data described in Table \ref{tab:data}.
The main purpose of our different setups is to test the capability of multilingual models not only in inside-language evaluation but also generalizing knowledge from one language to another. The generalization capability of offensive language detection models across different languages has been poorly explored.

For monolingual evaluation, we trained our model using each one of the training sets and the corresponding validation and testing sets. This experimental setup allows us not only to test the model in a specific language but also to obtain reference values to be compared to the ones obtained in the multilingual experimental setups. We refer to this setting as \emph{BERT Lang} (\emph{BERT Greek}, \emph{BERT English}, etc).

As a second configuration, we opted for multilingual training. We trained our model with the concatenation of all the training sets and evaluated it over each test set. The purpose of this experiment is to find languages that contribute positively to the monolingual classification. We call this setting \emph{BERT All}.

It might be argued that in monolingual settings it would have been a better option to simply use the \textit{BERT} version trained specifically for that language. However, we decided to use the multilingual version to have comparable results.

To assess the multilingual potential of BERT for this task, we also performed some \textit{zero-shot} and \textit{few-shot} experiments. Zero-shot experiments consisted in training the model in one language and evaluating it in a different one. That is, training with language \textit{A} and testing with language \textit{B}. Few shot experiments, on the other hand, trains the model in language \textit{A} using also a little amount of instances from \textit{B}, and tests their performance against \textit{B}. This cross-lingual generalization is desirable to tackle the same problem in low-resourced languages. In Section 5 we discuss the results in each case.

The evaluation metric proposed for this task is \emph{Macro F1}.

\subsection{Error Analysis and Model Interpretation}

To analyze the reasons behind the errors of our model, we used the Captum library \cite{captum2019github} implementation of Integrated Gradients \cite{sundararajan2017axiomatic} to have more information about the importance of each token towards classifying the tweet as offensive. This returns, for a model and a sentence, a value for each token representing the contribution of it towards the positive class (offensive) or towards the negative class (not offensive).

\section{Results}
\begin{table*}[t]
    \centering
    \begin{tabular}{l|lllll}
        Model & English & Danish & Greek & Arabic & Turkish \\ \hline
        Baseline & 0.720
  & 0.690& 0.639 & 0.547 & 0.452 \\
        %BERT English & \textbf{0.911} & 0.684          & 0.457          & 0.447         & 0.459 \\
        BERT English    & \textbf{0.895}          & 0.646          & 0.573          & 0.473         & 0.555 \\
        BERT Danish  & 0.713          & \textbf{0.740} & 0.473          & 0.473         & 0.463 \\
        BERT Greek   & 0.436          & 0.499         & \textbf{0.840} & 0.446         & 0.498 \\

        BERT Arabic  & 0.432          & 0.486          & 0.502          & \textbf{0.859}& 0.490 \\

        BERT Turkish & 0.431          & 0.501          & 0.499          & 0.545         & \textbf{0.766} \\
        BERT All     & \textbf{0.899} & \textbf{0.772} & \textbf{0.815} & \textbf{0.840}& \textbf{0.773} \\
        \hline \\
        Best System  & 0.922          & 0.812          & 0.852          & 0.901         & 0.825 \\
        Approx diff  & 0.023          & 0.040          & 0.037          & 0.061         & 0.052 \\
    \end{tabular}
    \caption{Performance of each model for the different datasets.  BERT Lang refers to BERT using that language as training data.
    %BERT Distant is BERT trained on the distant dataset. BERT All is the BERT multilingual model trained on all the data.
    The Baseline model is a bidirectional LSTM trained and tested in the same language. Performance is measured in Macro-F1 score. Non-diagonal entries show zero-shot performances. Bold letters indicate top two results.}
    \label{tab:results}
\end{table*}

Table \ref{tab:results} shows the results for each of our trained classifiers. The classifier presented for the competence is \emph{BERT All}, in spite of having better performing systems in the monolingual settings; for instance, \emph{BERT Greek} has better results for Greek than \emph{BERT All}.
However, \emph{BERT All} performs fairly well, having small differences with the best performing system for each language. For most languages, it stays in the ``top cluster'' of the competition for each language, most notably in Danish achieving the 7th position. We must remember that \textit{BERT All} is a single model for all the languages, reducing the need for several models -- in our case, a reduction of five-to-one.

% - 24° Arabic
% - 7° Danish
% - English, just 0.02 in Macro-F1 away from the top positions
% - 12 greek (0.04 from top)
% - Turkish 12 (0.05 from top)

In most cases, \emph{BERT All} showed results equal or slightly worse than the monolingual setting, telling us (at first sight) that adding other languages does not contribute to the overall performance. In the case the Turkish language, however, there is a slight increase from 0.766 to 0.773 of F-score. More interesting is the case of the Danish dataset, where the F-score increased from 0.74 to 0.77.

To find out the impact of the other languages over the Danish results, we conducted a data augmentation experiment. We augmented the training Danish dataset with other languages data to classify the tweets in the Danish dataset. Table \ref{tab:augmented_data} shows the results of this experiment. The results are, in general reasonable. Adding data from different languages does not dramatically impact the monolingual results. The addition of the Arabic dataset turned out to be the most successful one, despite having been the worst result in the zero-shot experiment. It is somehow surprising that a language from a very different family might positively impact in the performance of the Danish classifier.

\begin{table}
    \centering
    \begin{tabular}{l|l}
        Model &  Danish \\ \hline
        %BERT (Danish + Distant) & 0.720  \\
        %BERT (Danish + OLID) & 0.733   \\
        BERT (Danish + English) & 0.733   \\
        BERT (Danish + Greek) & 0.720  \\
        BERT (Danish + Turkish) & 0.697  \\
        BERT (Danish + Arabic)    & \textbf{0.792} \\
    \end{tabular}
    \caption{Performance on data augmentation experiments over the Danish test set. BERT Lang refers to BERT using that lang as training data. %BERT Distant is BERT trained on the distant dataset.
    Performance is measured in Macro-F1 score.}    \label{tab:augmented_data}
\end{table}

%
% Zero-shot
%

Regarding the zero-shot cross-lingual capability (the non-diagonal entries in Table \ref{tab:results}) it can be observed that there is no transfer learning in this mode. That is, no classifier trained in one language performs successfully when tested in a different one. The only exception for this is Danish-English, as we can see that training with English and testing against Danish gives something around 0.65 Macro F1. However, a close examination of the true positives in this case brings us to the conclusion that the transference is mainly due to vocabulary sharing and code switching.

%To analyze whether the classifier was unable to understand the offensiveness in Danish (perhaps a particular form of it in that language) we tested \emph{BERT English} against the English-translated version of the Danish dataset. In this case, however, the classifier performs well, achieving <NUM> Macro-F1.

\begin{figure}[t]
    \centering
    \includegraphics[scale=0.5]{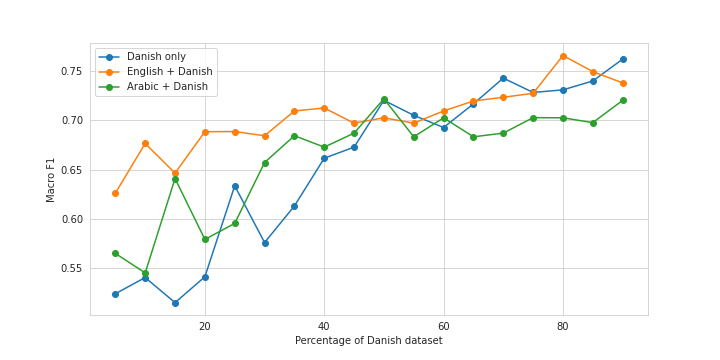}
    \caption{Results of Few-shot training for Danish. Blue line shows the performance of the classifier trained with the respective percentage of the training dataset, and the orange line corresponds to the classifier trained with English plus the respective percentage of the Danish dataset.}
    \label{fig:few_shot_danish}
\end{figure}

Few-shot experiments yielded slightly better results. Figure \ref{fig:few_shot_danish} displays the performance of classifiers trained with $5\%, 10\%, 15\% \ldots$ of the instances of the Danish dataset (blue line) and the performance of training the same classifier but using also another dataset: English or Arabic. Performance using $20\%$ of the Danish data and the English dataset achieves $0.69$ Macro-F1 score while we need almost the double of data to have the same performance using Danish-only training. Using Arabic dataset yields marginal improvement in performance.

\subsection{Error Analysis}

\begin{figure*}[t]
    \centering
    \begin{subfigure}{.45\textwidth}
        \centering
        \includegraphics[height=5cm]{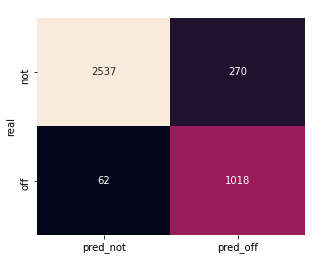}
        \caption{Confusion Matrix of BERT All in English}
        \label{fig:confusion_matrix}
    \end{subfigure}%
    \begin{subfigure}{.45\textwidth}
        \centering
        \includegraphics[height=5cm]{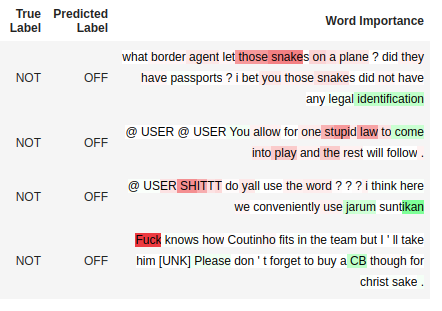}
        \caption{Word importance calculated by Integrated Gradients}
        \label{fig:word_importance}
    \end{subfigure}

    \caption{Error Analysis for our model. Figure \subref{fig:confusion_matrix} displays the confusion matrix of BERT All in English. Figure \protect{\subref{fig:word_importance}} displays the word importance for each token: red means that the token is pushing towards the ``offensive'' class, whereas green pulls to not offensive class. }
\end{figure*}

Figure \ref{fig:confusion_matrix} displays the confusion matrix for our model against the English test set. We can observe that most errors come from false positives. Figure \ref{fig:word_importance} shows the word importance for a couple of false positive examples. It turns out that numerous spurious correlations are learned by our classifier: words such as  ``Trump'', ``disgusting'', ``racist'' trigger the sentence as offensive. Also, more complex constructions such as reporting offensive incidents are not understood by the classifier: for instance, ``They call me b*tch'' should not be marked as offensive.

It is also important to notice that there are a number of instances that seem to be mislabelled. For instance, these examples were labelled as non-offensive:

\begin{itemize}
    \item what border agent let those snakes on a plane? did they have passports? i bet you those snakes did not have any legal identification
    \item It's Me and IDGAF About Nothing -- Females with nigga mindsets are dangerous
    \item Fuck knows how Coutinho fits in the team but I'll take him Please don't forget to buy a CB though for christ sake.
\end{itemize}

Whereas these are considered offensive:

\begin{itemize}
    \item Crazy that as we get older and go through certain shit you just want to keep it to yourself.
    \item @USER Trump lied..and then coldly said it now appeared his building was the tallest in NYC. He's a sick, sick twist.
    \item Knew sis was a liar but I got soft anyways smh
    \item It’s crazy how people make excuses for them to walk out of a person’s life...
\end{itemize}

In some cases it seems that there are instances that are wrongly labelled, and in other cases there are inconsistencies regarding swearing -- is saying ``sh*t'' or ``f*ck'' considered offensive? It's known that offensive language (alongside with cyberbullying, hate speech and so on) are difficult tasks for the annotators and that non-skilled annotators deliver datasets having a lot of noise.

\section{Conclusion}

In this work, we explored the capabilities of Multilingual BERT for offensive language detection in a multilingual scenario. We evaluated our model in different experimental setups: training and testing it individually for each language; training with one language data and testing in others (zero-shot mode); and training a single model on all languages.

Something to notice is that, in spite of the good performance of multilingual BERT in zero-shot cross-lingual evaluation for other tasks, it did not work well for this one. The only exception is the English-Danish cross-lingual evaluation, and it is mainly due to vocabulary intersection. Further work is needed to analyze the reasons why these zero-shot experiments failed. Nonetheless, the only experiment performed in few-shot mode showed slightly better results, letting a Danish model with just a hundred of examples achieve a reasonable performance.

%Add something about our few-shot experiments

On the other hand, training all the languages jointly resulted in a single model having a similar performance overall. This is interesting as having a single model instead of five has practical implications, in particular concerning the resources needed for big models such as BERT. 

The relationship of offensive language among different languages is something to be studied in more depth. At first glance, our experiments showed no zero-shot transference; a few-shot experiment showed (in the case of two languages of similar typology) some transference. Further study is needed to explain these results. 

\section*{Acknowledgements}

This material was supported by the Millennium Institute for Foundational Research on Data (IMFD) and by a research grant from SeCyT, Universidad Nacional de Cordoba.\\

\bibliographystyle{coling}
\bibliography{offenseval}

\end{document}